\pdfoutput=1
\documentclass[11pt,a4paper]{article}
\usepackage[hyperref]{naaclhlt2019}
\usepackage{times}
\usepackage{latexsym}
\usepackage{subcaption}
\usepackage{enumitem}
\usepackage{amsmath}
\usepackage{graphicx}
\usepackage{tikz}
\usepackage{caption}
\usepackage{pgfplots}
\usepackage{makecell}
\usepackage{booktabs, multirow, tabularx} 
\setlist[enumerate]{label*=\arabic*.}

\usepackage{url}

\aclfinalcopy 

\title{Nomic Embed Vision: Expanding the Latent Space}

\author{
    Zach Nussbaum\\
  {\tt zach@nomic.ai} \\ \And
    Brandon Duderstadt \\
  {\tt brandon@nomic.ai} \\ \And
      Andriy Mulyar \\
  {\tt andriy@nomic.ai} \\ 
}

\begin{document}

\maketitle

\begin{abstract}
This technical report describes the training
of nomic-embed-vision, a highly performant, open-code, open-weights image embedding model that shares the same latent space as nomic-embed-text. Together, nomic-embed-vision and nomic-embed-text form the first unified latent space to achieve high performance across vision, language, and multimodal tasks.
\end{abstract}


\section{Introduction}

Beginning with CLIP \cite{radford2021learning} and ALIGN \cite{jia2021scaling}, unsupervised multimodal encoders trained on large amounts of noisy web crawled data have shown impressive zero-shot capabilities across retrieval and classification tasks. 
These self supervised models are competitive with, and sometimes outperform, supervised baselines.
However, these models are only optimized for multimodal tasks, and the text encoders perform poorly on text-only benchmarks like MTEB \cite{muennighoff2023mteb, koukounas2024jina}.

Recently, Jina CLIP v1 \cite{koukounas2024jina} was introduced to address this issue. 
Unfortunately Jina CLIP does not achieve state of the art performance, failing to exceed jina-embeddings-v2 \cite{günther2024jina} on MTEB and OpenAI CLIP ViT B/16 \cite{radford2021learning} on Datacomp \cite{gadre2023datacomp} and Imagenet Zero-Shot Classification. 

In this technical report, we introduce nomic-embed-vision, a highly performant vision encoder that is aligned to the latent space of nomic-embed-text.
To train nomic-embed-vision, we adopt a similar training style to Locked Image Tuning (LiT) \cite{zhai2022lit}, but instead freeze a high-performing text embedder and train a vision encoder from a pretrained checkpoint.
This enables us to maintain the performance of nomic-embed-text as well as unlock new multimodal latent space capabilities.
Together, nomic-embed-vision and nomic-embed-text form the first unified latent space to achieve high performance across vision, language, and multimodal tasks.

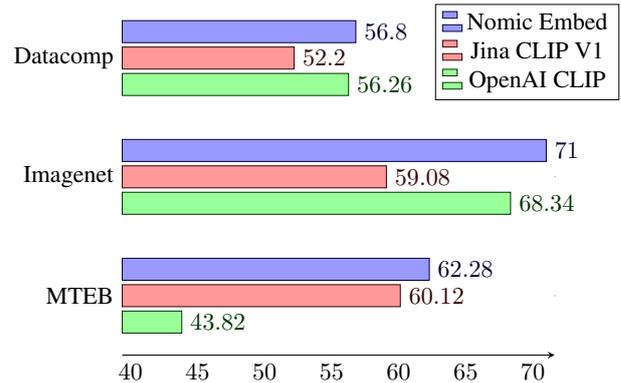
\begin{figure}
\begin{tikzpicture}[scale=0.83]
\begin{axis}[
xbar,
y axis line style = { opacity = 0 },
axis x line       = bottom,
reverse legend,
xmin = 40,
xlabel style={below},
xlabel={},
tickwidth         = 0pt,
enlarge y limits  = 0.25,
enlarge x limits  = 0.02,
symbolic y coords = {MTEB,Imagenet, Datacomp},
legend style={at={(1.15,0.85)},anchor=east},
nodes near coords,
]

\addplot[green!20!black,fill=green!40!white] coordinates {(43.82,MTEB) (56.26,Datacomp) (68.34,Imagenet)  }; 
\addplot[red!20!black,fill=red!40!white] coordinates {(60.12,MTEB) (52.20,Datacomp) (59.08,Imagenet)  }; 
\addplot[blue!20!black,fill=blue!40!white] coordinates {(62.28,MTEB) (56.8,Datacomp) (71.00,Imagenet)    };
\legend{OpenAI CLIP, Jina CLIP V1, Nomic Embed}
\end{axis}
\end{tikzpicture}
\caption
{\textbf{Multimodal and Text Embedding Benchmark} Aggregate performance of Nomic Embed v1.5, OpenAI CLIP ViT B/16, and Jina CLIP v1 on text and multimodal benchmarks. Nomic Embed V1.5 is the only multimodal encoder to outperform OpenAI CLIP on multimodal and text benchmarks.
X-axis units vary per benchmark suite. Imagenet is Imagenet Zero-Shot, Datacomp is a suite of 38 zero-shot multimodal evaluations, and MTEB evaluates performance of text embedding models.}
\label{multimodal}
\end{figure}

\section{Related Work}
Large scale noisy contrastive pretraining of image and text encoders was pioneered by \citet{radford2021learning, jia2021scaling} using a large batch size and InfoNCE loss \cite{oord2019representation}.

CLIP-style models are trained across a large noisy dataset created by crawling the web and extracting image-text pairs from webpages.
These models are generally trained on billions of image-text pairs with a large batch size, which results in a massive pretraining compute requirement.

\citet{radford2021learning} originally proposed evaluating CLIP models using zero-shot accuracy across 27 datasets.
Unfortunately, the lack of public information regarding the composition of the original web scale train set complicates this evaluation. 
To remedy this, \citet{gadre2023datacomp} introduced Datacomp, an open benchmark to evaluate both CLIP-style models and their constituent training data mixes.

Taking inspiration from transfer learning, LiT \cite{zhai2022lit} and aligns a text encoder to a frozen pretrained image encoder, reducing the compute required to train a quality multimodal encoder.
Three Towers \cite{kossen2023towers} improved upon LiT by introducing a third frozen pretrained image encoder and allowing the image and text encoders to take advantage of contrastive training as well as pretrained embeddings.

Imagebind \cite{girdhar2023imagebind} learns a joint embedding across many modalities by aligning modalities (e.g. audio) utilizing only image-paired data starting with a ViT-H from OpenCLIP \cite{ilharco_gabriel_2021_5143773}.

\begin{table*}[htb]
    \centering
\begin{tabular}{lcccccc}
  \toprule
  Vision Encoder & Pretrain & Supervised & IN-ZS & $I->T$ & $T->I$ & Mean R@1 \\
  \midrule
  Randomly Initialized & N/A & N/A & 41.20 & 35.50 & 28.48 & 31.99  \\
  \href{https://huggingface.co/google/vit-base-patch16-224}{ViT} \cite{dosovitskiy2021image} & IN21k & Y & 62.64 & 49.60 & 40.32 & 44.96 \\
  \href{https://huggingface.co/timm/vit_base_patch16_224.augreg_in21k}{AugReg} \cite{steiner2022train} & IN21k & Y & 57.56 & 50.80 & 42.88 & 46.84 \\
  \href{https://huggingface.co/timm/vit_base_patch16_rope_reg1_gap_256.sbb_in1k}{ViT RoPE} \cite{rw2019timm} & IN1k & Y & 61.25 & 52.50 & 42.32 & 47.41 \\ 
  \href{https://huggingface.co/timm/eva02_base_patch14_224.mim_in22k}{Eva02} \cite{fang2023eva02} & IN21K & N & \textbf{65.19} & \textbf{59.90} & \textbf{48.32} & \textbf{54.11} \\
  \bottomrule
\end{tabular}
    \caption{Effect of initialization of vision backbone on Imagenet Zero-shot and Flickr 30k Image to Text Recall@1, Text to Image Recall@1, and mean Recall@1. The pretrain column refers to the dataset the vision encoder was pretrained on and Supervised is whether the vision encoder used a supervised task to pretrain.}
    \label{tab:backbone}
\end{table*}

Text embedding models are similarly trained contrastively on a large collection of text pairs and initializing with a pretrained transformer. \citet{reimers2019sentencebert} train a pretrained BERT model contrastively for sentence similarity tasks. Since then, models such as E5 \cite{wang2024text}, GTE
\cite{li2023general}, BGE \cite{xiao2024cpack}, InstructOR \cite{su2023embedder}, Jina \cite{günther2024jina}, and Nomic \cite{nussbaum2024nomic} train dual encoders in multiple stages. 

MTEB \cite{muennighoff2023mteb} aims to evaluate text embedding models across a suite of tasks including classification, retrieval, and semantic similarity. 

\begin{figure}[t]
    \centering
    \includegraphics[width=\linewidth]{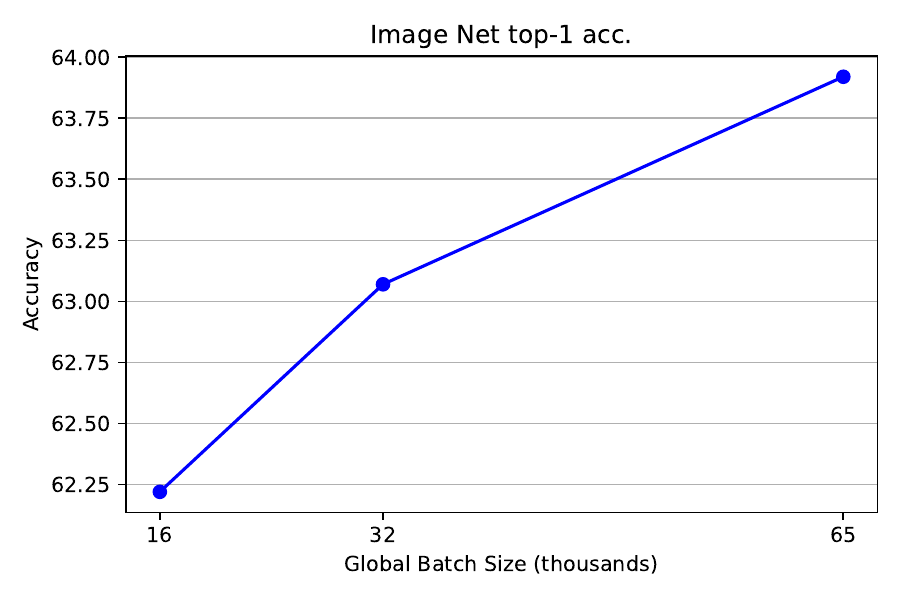}
    \caption{Imagenet Zero-Shot Top 1 Accuracy improves as we increase batch size in small scale experiments}
    \label{fig:batch_size}
\end{figure}

\section{Methods}

Our goal is to learn a unified embedding space that performs well on multimodal tasks as well as unimodal text and image tasks.
Contrastive Image Text Pretraining as introduced by \citet{radford2021learning} leads to high performing multimodal models \cite{radford2021learning,jia2021scaling}. However, as shown in Figure 1 and noted by \citet{koukounas2024jina}, training only on these large scale datasets leads to poor general text embedding performance. 
\subsection{Image Text Contrastive Training}
Training CLIP-style models from scratch is expensive and requires large amounts of compute and data.
\citet{zhai2022lit} investigated ways to train CLIP models in a more efficient manner by freezing a pretrained vision encoder and training the text encoder from scratch. This methodology, which they named LiT, extends any pretrained vision encoder multimodal and zero-shot capabilities.

However, one downside of LiT is that freezing the image encoder prevents the vision encoder's representations from being updated with signal from the text data.
To remedy this, \citet{kossen2023towers} proposes using a third frozen image tower to transfer representations to the main image and text encoders that are trained from scratch. 
This approach allows the encoders to be updated during training while also benefiting from the pretrained representations of the vision encoder. Three Towers outperforms LiT and CLIP-style models on retrieval tasks across initializations and pretraining datasets.

\citet{koukounas2024jina} proposes a three stage contrastive training strategy to learn multimodal and text representations. In the first stage, they train the image and text encoders, initializing from EVA02 \cite{fang2023eva02} and a pretrained JinaBERT model, similar to \cite{günther2024jina} and optimize the image-text and text-text alignment.
The second stage uses longer synthetic captions for further image-text alignment. The third stage introduces hard negatives to the text-text alignment to improve text embedding performance. 

Similarly to \cite{koukounas2024jina}, we aim to train general, high performing multimodal encoders. In this work, we adapt the LiT \cite{zhai2022lit} training recipe, and instead freeze the text encoder. 
Our early work in adapting Three Towers style \cite{kossen2023towers} training resulted in poor general text embedding models, so we focused our effort on Locked Text Tuning.

\section{Image Text Datasets}
\citet{radford2021learning} describes curating a dataset of 400 million image-text pairs by searching for images that overlap with 500,000 popular phrases. This dataset was never released publicly. 

Subsequent works by \citet{schuhmann2021laion400m} and \citet{schuhmann2022laion5b} openly released Laion 400M and Laion 5B to facilitate the training of open source multimodal models.
\citet{xu2024demystifying} aim to reproduce the data curated in \citet{radford2021learning} and outperforms the proprietary dataset without any reliance on an external model.

\citet{gadre2023datacomp} also released Datacomp 1B, a top performing dataset on the Datacomp X-Large benchmark. \citet{fang2023data} improves upon the dataset released in \cite{gadre2023datacomp} by learning a data filtering network that can be used to curate high quality image-text datasets. In this work, we use Data Filtering Networks 2B (DFN-2B), the curated dataset for the Datacomp X-Large track. At the time of curation, we were only able to obtain 1.5B of the 2B links. 

\section{Experiments}
Nomic Embed Vision v1 and v1.5 were trained with identical hyperparameters and recipies except for the initialization of their text encoders.
We train on DFN-2B for 3 epochs with a batch size of 65,536, resulting in training on ~5B samples.
We initialize the text encoders for Nomic Embed Vision v1 and v1.5 from Nomic Embed Text v1 and v1.5 respectively \cite{nussbaum2024nomic}, and the vision encoder as EVA02-ViT B/16 \cite{fang2023eva02}.
We use the AdamW optimizer \cite{loshchilov2019decoupled} and a peak learning rate of 1e-3, 2000 warmup steps, and cosine decay.
As noted in \citet{zhai2023sigmoid}, we set weight decay to 0 for the pretrained vision encoder. 
We employ multi-head attention pooling \cite{kossen2023towers, big_vision, rw2019timm}.
We train on 224x224 pixel images and use the same image preprocessing as \cite{radford2021learning}.
We additionally employ small augmentations using random crops \cite{ilharco_gabriel_2021_5143773} and do not clamp the learnable logit scale unlike \citet{radford2021learning}.

\subsection{Evaluation of Design Decisions}
Due to the high compute and time cost to training the full model, we explored different design decisions at smaller scales. We present evidence in favor of some of our design decisions.

For our small scale experiments, we train for 1 epoch and perform small hyperparameter searches over learning rate and weight decay. 
We employ the Locked Text Tuning strategy outlined above and freeze Nomic Embed Text v1.

\subsection{Evaluating Batch Size}
As noted in \cite{zhai2022lit,chen2020simple,radford2021learning}, large batch sizes can improve the performance of contrastively trained models. We initialize the vision encoder with a ViT B/16 from \cite{dosovitskiy2021image} and train on 300M image-text pairs over Data Filtering Networks from the Datacomp Large track (DFN-Large) \cite{fang2023data,gadre2023datacomp}.

As shown in Figure \ref{fig:batch_size}, increasing the batch size leads to sizable improvements on ImageNet 0-shot accuracy.
We choose to use 65,536 as this is the biggest batch size we can accomodate given our compute limitation.
We leave it to future work to investigate whether performance increases from increased batch size plateau. 

\subsection{Evaluating Pretrained Vision Encoders}
To evaluate pretrained visison encoders, we train for one epoch on DFN-Large \cite{fang2023data}.
For each encoder, we perform a sweep over learning rate and weight decay.
We investigate vision encoders released in \citet{dosovitskiy2021image}, \citet{steiner2022train}, \citet{rw2019timm}, and \citet{fang2023eva02}.

Similar to \cite{zhai2022lit}, we found that the pretrained vision encoder backbone had a large effect on the quality of the final model, particularly in regards to multimodal retrieval.
As shown in Table \ref{tab:backbone}, we find that more broadly pretrained vision encoders lead to better multimodal retrieval and Imagenet zero-shot results. For example, a supervised vision encoder like the ViT B/16 released in \cite{dosovitskiy2021image} performs well on Imagenet zero-shot but poorly on Flickr 30k retrieval \cite{young-etal-2014-image}.
Recently released ViTs from \cite{rw2019timm} using techniques like global registers \cite{darcet2023vision} and rotary positional embeddings \cite{fang2023eva02} show promise even though they are trained on a small dataset like Imagenet. 

From Table \ref{tab:backbone}, we notice that even though models released by \citet{rw2019timm} and \citet{steiner2022train} are trained in a supervised manner, they outperform the ViT B/16 released by \citet{dosovitskiy2021image}. 
As noted by \citet{zhai2022lit}, training on large amounts of data leads to better and more general visual representations, even when training without a supervised objective.
We hypothesize the high performance of the ViT B/16 released in \citet{rw2019timm} is due to using heavy augmentation, like RandAugment \cite{cubuk2019randaugment}, and training for many epochs. 

Ultimately, the ViT B/16 released in \cite{fang2023eva02} performed the best across Imagenet zero-shot and Flickr retrieval, which leverages the unsupervised Masked Image Modeling (MIM) objective \cite{bao2022beit}.

\begin{figure}[t]
    \centering
    \includegraphics[width=\linewidth]{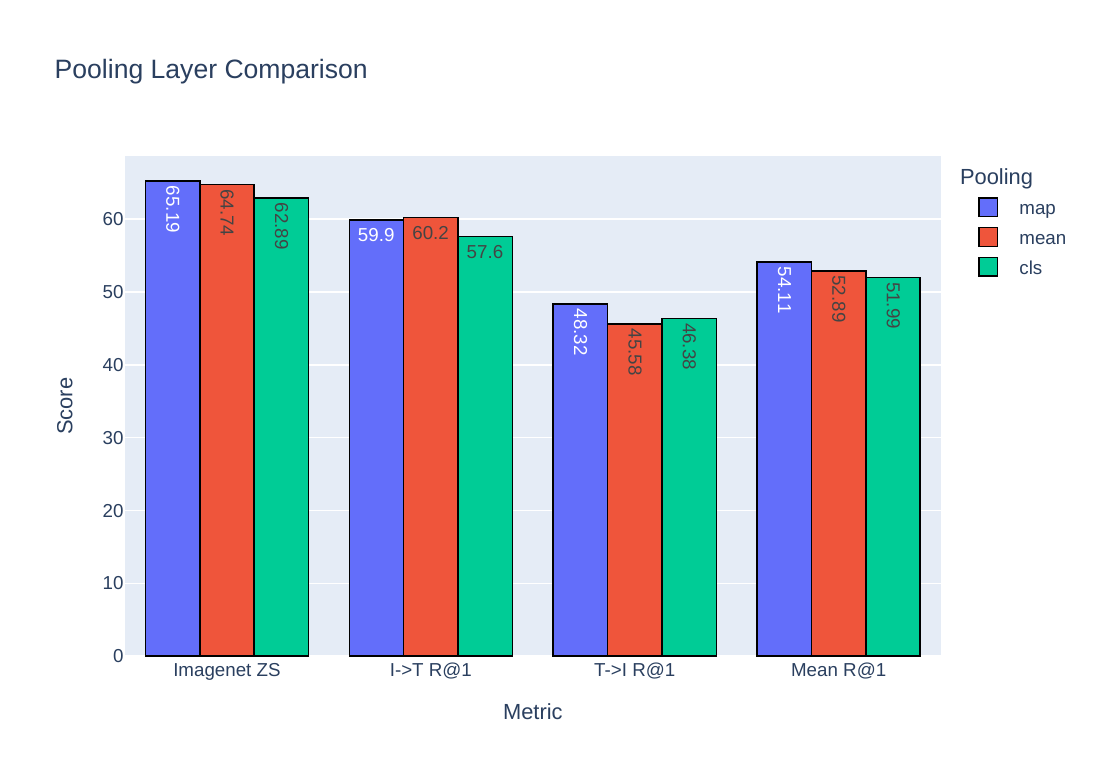}
    \caption{Effect of Pooling Layer on Performance in various retrieval and classification setups.}
    \label{fig:pooling}
\end{figure}
\subsection{Evaluating Pooling Strategies}
We evaluate different pooling layers for the vision encoder.
We compare using the class token pooling, mean pooling, and multihead attention pooling \cite{kossen2023towers,big_vision}. 
Again, our small scale experiments consist of training for 1 epoch over DFN Large \cite{fang2023data}.
We initialize the pretrained vision encoder from \citet{fang2023eva02}.
We find that mutlihead attention pooling (MAP) performed the best over Imagenet Zero-Shot and Flickr 30k as shown in Figure \ref{fig:pooling}.

\begin{table*}[htb]
    \centering
\begin{tabular}{|l|c|c|c|c|c|}
\hline
\textbf{Model} & \textbf{ImageNet} & \textbf{ImageNet dist. shifts} & \textbf{VTAB} & \textbf{Retrieval} & \textbf{Average} \\
\hline
Nomic Embed v1.5 & 0.710 & 0.551 & 0.561 & 0.469 & 0.568 \\
Nomic Embed v1 & 0.707 & 0.551 & 0.565 & 0.457 & 0.567 \\
CLIP ViT B-16 & 0.684 & 0.559 & 0.546 & 0.527 & 0.563 \\
Jina CLIP v1 & 0.591 & 0.464 & 0.520 & 0.604 & 0.522 \\ 
\hline
\end{tabular}
\caption{Model Performance on DataComp Classification and Retrieval Tasks}
\label{tab:model_performance}
\end{table*}

\section{Training Resources}
Nomic Embed Vision v1 and v1.5 were trained on 2 8xH100s over 3.5 days.
DFN-2B requires ~62TB to store and several thousand dollars to preprocess.

\section{Discussion}
We present a recipe for enhancing a high quality text embedder with multimodal capabilities.
While this model outperforms other unified embedding spaces, there are several important caveats.
Consistent with prior CLIP literature, Nomic Embed exhibits bag of words like behavior on some tasks. \cite{yuksekgonul2023visionlanguage,paiss2023teaching}. 
We also find that the retrieval scores resulting from Locked Text Tuning tend to skew low compared to similarly performing CLIP-style models as shown in Table \ref{tab:model_performance}.
Three towers training \cite{kossen2023towers} presents a promising direction for remedying this.

Future work can investigate if similar strategies to those shown in \citet{tschannen2023image} can be adapted with a strong general purpose text encoder. 
However, some modifications may have to be made as the text encoder used in this work is bidirectional. 

Moreover, recent work on multimodal embedding space geometry suggests that CLIP style training is not sufficient for closing the modality gap present in multimodal embedding spaces. \citet{liang2022mind, zhang2024connect}
As a result, we refer to the embedding spaces of Nomic Embed Text and Nomic Embed Image as unified and not aligned in this work.
We leave it to future work to investigate whether closing the modality gap improves downstream performance.

\section{Conclusion}
We adapt the contrastive tuning framework presented in \cite{zhai2022lit} to enhance a high performing text encoder with multimodal capabilities.
We call this training paradigm Locked Text Tuning, and use it to train Nomic Embed Vision.
Together, Nomic Embed Vision and Nomic Embed Text form the first unified latent space to achieve high
performance across vision, language, and multimodal tasks.

\bibliographystyle{acl_natbib}
\bibliography{nomicembed}

\subsection*{Appendix}
\vspace{-10cm}
\begin{table*}[htb]
    \centering
    \setlength{\tabcolsep}{4.5pt} 
    \caption{Detailed performance on the CLIP Benchmark. Numbers for JinaCLIP \cite{koukounas2024jina}, OpenAI CLIP \cite{radford2021learning}, EVa02-CLIP \cite{fang2023eva02}, and Long CLIP \cite{zhang2024longclip} reported from \citet{koukounas2024jina}}
    \label{appendiX:clip-benchmark}
    \vspace{0.1in}
    \begin{center}
    \begin{small}
    \begin{tabular}{l|ccccc}
        \toprule
        Model & \makecell{JinaCLIP} & \makecell{Nomic Embed} & \makecell{OpenAI CLIP } & \makecell{EVA02-CLIP} & \makecell{LongCLIP} \\
        \midrule
        \multicolumn{6}{c}{\textbf{Zero-shot Image Retrieval - Recall@5  [\%]}} \\
        \midrule
        Average & 80.31 & 69.43 & 75.62 & \textbf{82.15} & 81.72 \\
        Flickr30k & 89.02 & 77.98 & 85.60 & \textbf{91.10} & 90.46 \\
        Flickr8k & 85.50 & 74.10 & 82.84 & \textbf{88.50} & 88.40 \\
        MSCOCO  & 66.42 & 56.21 & 58.42 & \textbf{66.85} & 66.31 \\
        \midrule
        \multicolumn{6}{c}{\textbf{Zero-shot Text Retrieval - Recall@5  [\%]}} \\
        \midrule
        Average & 89.91 & 80.44 & 88.12 & 90.59 & \textbf{90.79} \\
        Flickr30k  & 96.50 & 89.89 & 96.20 & 96.60 & \textbf{98.00} \\
        Flickr8k & 94.20 & 84.50 & 91.40 & \textbf{94.60} & 94.00 \\
        MSCOCO  &  79.02 & 67.02 & 76.76 & \textbf{80.58} & 80.38 \\
        \midrule
        \multicolumn{6}{c}{\textbf{Image Classification - Accuracy@1 [\%]}} \\
        \midrule
        Average & 43.28 & 46.62 & 46.16 & \textbf{48.70} & 46.67 \\
        Cars & 68.03 & \textbf{87.60} & 64.73 & 78.56 & 59.17 \\
        Country211 & 13.45 & 16.35 & \textbf{22.85} & 21.34 & 20.28 \\
        Fer2013 & \textbf{49.07} & 20.30 & 46.18 & 51.17 & 47.80 \\
        Fgvc-aircraft & 11.49 & 23.64 & 24.27 &\textbf{ 25.11} & 22.56 \\
        Gtsrb & 38.70 & 45.22 & 43.58 & \textbf{46.33} & 42.93 \\
        Imagenet-a & 29.92 & 46.04 & 49.93 & \textbf{53.89} & 46.84 \\
        Imagenet-o & 33.40 & 20.55 & \textbf{}42.25 & 34.10 & \textbf{42.65} \\
        Imagenet-r & 73.66 & \textbf{82.46} & 77.69 & 82.42 & 76.63 \\
        Imagenet1k & 59.08 & 71.03 & 68.32 & \textbf{74.75} & 66.84 \\
        Imagenet-sketch & 45.04 & 57.51 & 48.25 & \textbf{57.70} & 47.12 \\
        Imagenetv2 & 51.37 & 62.17 & 61.95 & \textbf{66.98} & 60.17 \\
        Mnist & 48.07 & 59.42 & 65.51 & 47.16 & \textbf{71.84} \\
        Objectnet & 45.41 & 62.02 & 55.35 & \textbf{62.29} & 50.79 \\
        Renderedsst2 & 59.14 & 55.29 & \textbf{60.68} & 54.15 & 59.31 \\
        Stl10 & 97.89 & 97.47 & 98.28 & \textbf{99.49} & 98.41 \\
        Sun397 & 65.92 & 65.12 & 64.37 & \textbf{70.62} & 68.73 \\
        Voc2007 & 72.83 & 61.75 & 78.34 & \textbf{80.17 }& 75.35 \\
        Vtab/caltech101 & 82.68 & \textbf{84.58} & 82.19 & 82.78 & 82.63 \\
        Vtab/cifar10 & 93.49 & 96.82 & 90.78 & \textbf{98.46} & 91.22 \\
        Vtab/cifar100 & 72.08 & 83.62 & 66.94 & \textbf{87.72} & 69.17 \\
        Vtab/clevr-closest-object-distance & 15.61 & 15.84& 15.83 & 15.72 & \textbf{15.90} \\
        Vtab/clevr-count-all & \textbf{22.35} & 21.62 & 21.09 & 21.27 & 20.71 \\
        Vtab/diabetic-retinopathy & 2.82 & 4.51 & 3.44 & 14.19 & 10.99 \\
        Vtab/dmlab & \textbf{19.53} & 13.97 & 15.49 & 14.67 & 15.45 \\
        Vtab/dsprites-label-orientation & 2.44 & 1.63 & 2.34 & 1.94 & 1.12 \\
        Vtab/dsprites-label-x-position & 3.07 & 2.95 & 2.95 & 3.11 & \textbf{3.15} \\
        Vtab/dsprites-label-y-position & 3.17 & 2.87 & 3.11 & \textbf{3.21} & 3.16 \\
        Vtab/dtd & \textbf{55.43} & 50.27 & 44.89 & 52.82 & 45.27 \\
        Vtab/eurosat & 49.52 & 37.27 & 55.93 & \textbf{66.33} & 60.44 \\
        Vtab/flowers & 59.62 & 68.23 & 71.13 & \textbf{75.75} & 69.85 \\
        Vtab/kitti-closest-vehicle-distance & 22.93 & \textbf{38.82} & 26.44 & 22.08 & 34.60 \\
        Vtab/pcam & 55.54 & 61.48 & 50.72 & 50.95 & 52.55 \\
        Vtab/pets & 80.98 & 91.79 & 89.04 & \textbf{92.10} & 89.21 \\
        Vtab/resisc45 & 55.46 & 57.12 & 58.27 & 60.37 & \textbf{60.63} \\
        Vtab/smallnorb-label-azimuth & \textbf{5.40} & 5.30 & 5.21 & 4.96 & 5.14 \\
        Vtab/smallnorb-label-elevation & 11.31 & 9.62 & \textbf{12.17} & 9.79 & 10.59 \\
        Vtab/svhn & 25.46 & \textbf{42.70} & 31.20 & 17.65 & 27.65 \\
        \bottomrule
    \end{tabular}
\end{small}
\end{center}
\end{table*}

\end{document}